\documentclass[11pt,a4paper]{IEEEtran}
\usepackage[latin1]{inputenc}
\usepackage[english]{babel}
\usepackage[T1]{fontenc}
\usepackage{array}
\usepackage{graphicx}
\usepackage[caption=false,font=footnotesize]{subfig}
\usepackage{cite}
\usepackage{algorithm}
\usepackage[noend]{algpseudocode}
\usepackage{subfig}
\usepackage{wrapfig}
\usepackage{transparent}
\usepackage[cmex10]{amsmath}
\usepackage{amsfonts,amssymb}
\usepackage{bm}
\usepackage{booktabs}
\usepackage{color}
\usepackage{epstopdf}
\usepackage{float}
\usepackage{multicol}
\usepackage{soul}
\usepackage{url}
\usepackage{indentfirst}
\usepackage{tikz}
\definecolor{gold}{rgb}{0.85,.66,0}
\newcommand{\ta}{\textcolor{blue}}

\DeclareMathAlphabet\mathbfcal{OMS}{cmsy}{b}{n}

\usepackage{xspace,colortbl}

\begin{document}

\title{Throughput and Latency in the Distributed Q-Learning Random Access mMTC Networks}
\author{{Giovanni~Maciel~Ferreira~Silva}, {Taufik~Abr\~ao}\\
\thanks{G. Maciel and T. Abr\~ao are with Department of Electrical Engineering, State University of Londrina, Parana, Brazil. E-mail: giomaciel.fs@gmail.com,\,\, taufik@uel.br}}
\date{\today}

\maketitle

\begin{abstract}
In mMTC mode, with thousands of devices trying to access network resources sporadically, the problem of random access (RA) and collisions between devices that select the same resources becomes crucial. A promising approach to solve such an RA problem is to use learning mechanisms, especially the Q-learning algorithm, where the devices learn about the {best time-slot periods} to transmit through rewards sent by the central node. In this work, we propose a distributed packet-based learning method by varying the reward from the central node that favors devices having a larger number of remaining packets to transmit. Our numerical results indicated that the proposed distributed packet-based Q-learning method attains a much better throughput-latency trade-off than the alternative independent and collaborative techniques in practical scenarios of interest. In contrast, the number of payload bits of the packet-based technique is reduced regarding the collaborative Q-learning RA technique for achieving the same normalized throughput. 	
\end{abstract}

\section{Introduction}\label{sec:intro}
Since the beginning of studies on the fifth generation of wireless communications (5G), it is known that the paradigm is not simply to increase transmission rates \cite{Popovski2018_NetworkSlicing}. With the demand for services such as the Internet of Things (IoT), smart cities, smart homes, among others, we seek to solve the problem of ensuring connectivity for thousands of devices at an access point. Because of this, 5G has been divided into three main modes: enhanced mobile broadband (eMBB), to guarantee high rates for mobile users; ultra reliable and low latency communications (URLLC), to guarantee a low latency and high reliability connection to certain services such as remote surgery, and massive machine-type communications (mMTC), to connect thousands of machine-type devices to the network.

In mMTC mode, thousands or even tens of thousands machine-type devices access network resources sporadically, i.e., a sensor network that sends data every minute. In this way, the transmission rate is not one of the most important figures of merit, but throughput, which evaluates the number of successful transmissions in a given time interval, and the probability of collision, which is the number of collisions that occurs at a certain time interval \cite{Bockelmann2016_mMTC5g,Bockelmann2018_ScalableMMTC}.

The problem that arises in this mode is to solve the random access (RA), since the devices randomly select the time-slot to transmit and collisions of two or more devices may occur, making communication impossible. Traditionally, the slotted ALOHA (SA) is used to solve the RA problem. In this technique, colliding devices retransmit after a fixed time-slot window, which reduces the probability of a new collision. However, it has been shown \cite{Sharma2019_CongestionMinimization} that SA has a high probability of collision in highly congested scenarios, where the number of devices is greater than or equal to the number of time-slots in a frame. Many solutions to solve the RA problem in mMTC are present in the literature, based on several different techniques, such as the grant-free ALOHA \cite{Qi2020_ALOHAGrantFree_RandomAccess}, the unequal access latency (UAL) \cite{Jiao2020_UAL_RandomAccess}, the sparse signal recovery \cite{Cui2020_SparseSignap_DeepLearning}, and the distributed queue-based framework \cite{Bui2019_RandomAccess_QueueBased}.

An alternative and promising way to solve the RA problem is the use of machine learning (ML) techniques \cite{Sharma2019_survey}, where the devices themselves learn to choose the best time-slots to transmit, avoiding collisions and increasing throughput. Q-Learning, being model-free, is a viable solution in these scenarios since machine-type devices must carry out learning in a simplified and not very complex way. 
Machine learning-based techniques will play an important role in technologies foreseen for the sixth generation of wireless communication (6G) \cite{Chowdhury2020_6G_Applications, Zhang2019_6G_Visions, Khan2020_6G_FutureDirections}.

Recently, techniques based on Q-Learning are present in many works in the field of wireless communications. Some applications include network slicing \cite{Li2020_QLearning_Slicing}, spectrum access \cite{Su2020_QLearning_SpectrumAccess}, non-orthogonal multiple access (NOMA) \cite{Valente2020_QLearning_NOMA}, and geographic routing for unmanned robot networks (URNs) \cite{Jin2019_QLearning_Reward}.

One of the simplest ways to use Q-Learning to {mitigate the collision problem in RA protocols} is the independent technique, where central node sends a binary reward to the devices, informing if the transmission was successful or if there was a collision between two or more devices. {Such} technique does not perform well in scenarios where the number of devices is equal to or greater {than the} number of available time-slots. In contrast to it, the collaborative Q-Learning  method is suggested, where the reward sent to the colliding devices is the level of congestion in the time-slot. In this case, the devices learn to choose the least disputed time-slots, increasing the throughput of the system \cite{Sharma2019_CongestionMinimization}.

An alternative to the independent {Q-Learning} technique is the collaborative approach, which considers the level of congestion in the reward sent from the central node to the devices. The throughput is higher for this technique than the independent technique, however the reward needs to be sent in more than one bit and the central node needs to know the number of devices that collided in a given time-slot.

Both techniques mentioned are not fair, as the devices that randomly select the least disputed time-slots will transmit their packages more quickly, as the learning method will provide them with unique time-slots. On the other hand, devices that collide frequently will take longer to transmit all of their packets.

The {\it Contribution} of this work is to propose a Q-Learning RA technique that does not detract from the devices that select the most congested time-slots at the beginning of the transmission, as occurs in the collaborative technique proposed in \cite{Sharma2019_CongestionMinimization}. The proposed distributed packet-based {Q-Learning} technique benefits devices that still have many packets to transmit, sending them a greater reward. The technique in \cite{Sharma2019_CongestionMinimization} sends larger rewards to devices that have uniquely selected time-slots, causing some devices to end transmission very quickly, while others take longer.  {In general, the} distributed packet-based method reduces the latency variance. {Also, we have proposed an improvement} in the collaborative {Q-Learning} technique aiming at {establishing a reasonable level of congestion with a finite number of bits, and as result, reducing} the header when sending the reward.

The remainder of the work is composed of the system model in Section \ref{sec:model}, the proposed distributed packet-based Q-Learning reward method in Section \ref{sec:qlearning}; numerical results {are analyzed} in Section \ref{sec:results}{; the main} conclusions and final remarks {are presented in} Section \ref{sec:conclusions}.

\section{System model}\label{sec:model}
{Let's consider} an mMTC network, where there are $N$ machine-type devices transmitting packets with $p$ bits of payload to a central node. A frame is made up of $K$ time-slots, and the $N$ devices select one of the slots to transmit. Each device has $L$ packets to transmit, with only one packet being transmitted per time-slot. The loading factor is given by the ratio between the number of active devices and the number of time-slots within a frame, $\mathcal{L}=\frac{N}{K}$.  The indexes for each device and each time-slot are sorted in sets $\mathcal{N} = \{1,\dots,N\}$ and $\mathcal{K} = \{1,\dots,K\}$, respectively. 
In addition, we define the {set $\psi_k$ indicating} which devices have chosen the $k$-th time-slot. For example, if devices 2 and 5 selected the 3rd time-slot, then $\psi_3 = \{2,5\}$.

The transmission of a packet is considered successful when only one device selects the $k$-th time-slot, {{\it i.e} it results in cardinality one, $|\psi_k| = 1$. Otherwise, if two or more devices choose the same time-slot, $|\psi_k| > 1$, a collision occurs,} and the transmission is considered a failure.

{For simplicity of analysis,} the effects of physical channel losses such as multipath fading and AWGN {(high SNR regime)} are not considered. As the focus of the work is on developing {the reward sending mechanisms in Q-Learning-based RA   protocols} aiming to improve the learning {process} of the devices, {we assume, as in \cite{Sharma2019_CongestionMinimization,Sharma2019_survey},} that the signal from all devices arrive with the same power at the central node. The random {variables are defined by} the channel{/slot} selection {in each device}. In addition, central node does not apply any collision resolution method to decide which device wins. When two or more devices select the same channel to transmit, central node considers it a collision and requests that the devices retransmit the packet.

{At} the end of the frame, central node sends a {{\it reward} signaling} to the devices {composed by} $b$ bits indicating whether the transmission was successful or not. The devices use the reward {information} to learn over the transmissions which are the best {time-slots subset} to transmit. The end of transmission occurs when all devices transmit all their packets.

\begin{figure}[!htbp]
\centering
\includegraphics[width=.85\columnwidth]{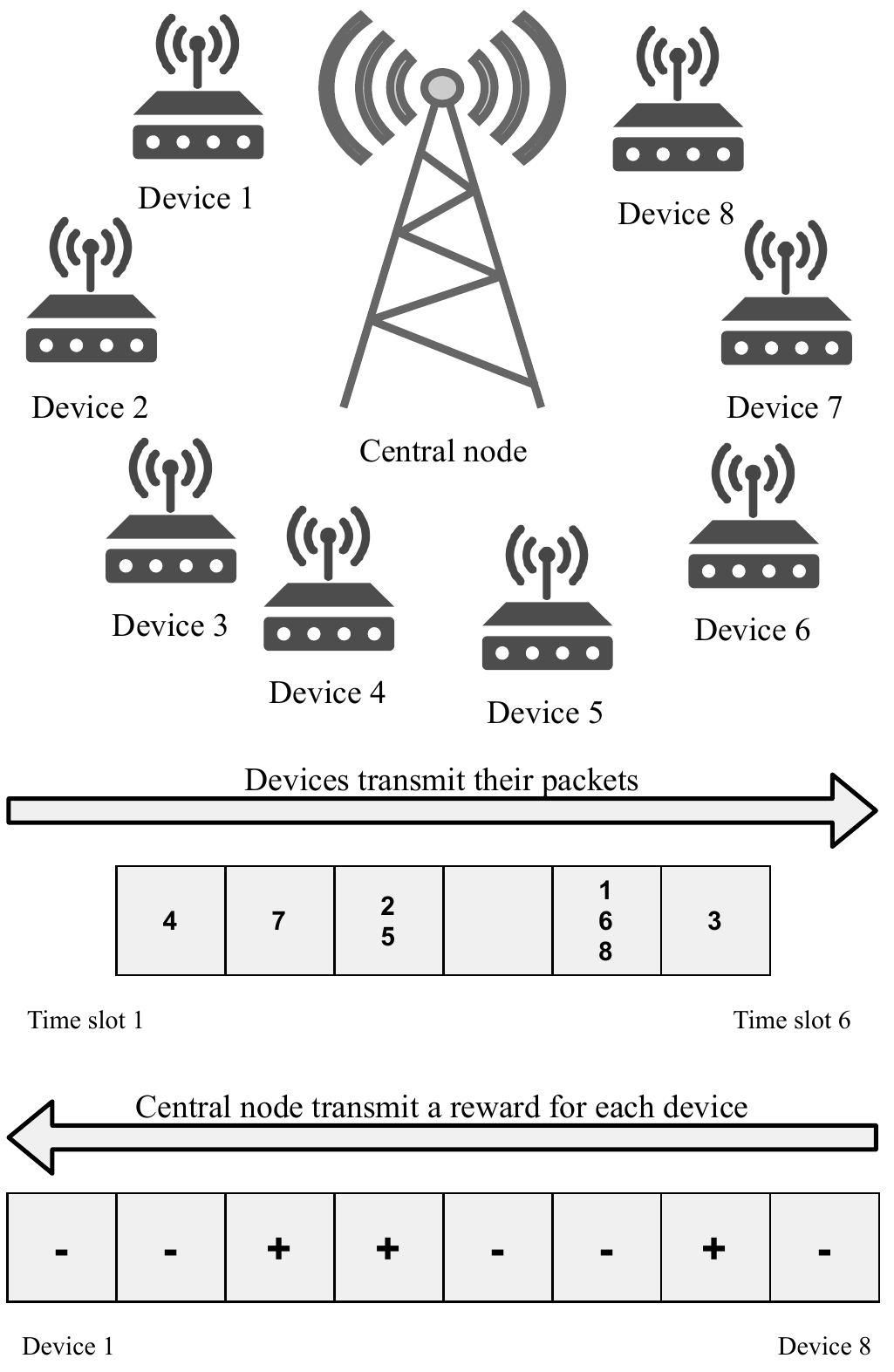}
    \caption{{Reward-based RA network with $N$ = 8 devices and $K$ = 6 time-slots.}}
    \label{fig:system_model}
\end{figure}

{To illustrate the transmission process across $K=6$ time-slots, Fig. \ref{fig:system_model} depicts the RA-based network} considering $N$ = 8 {devices.} In this {simple} example, only devices 3, 4 and 7 select time-slots exclusive to them. Therefore, they are the only ones to receive a positive reward from the central node. On the other hand, the other devices collide with each other, and therefore the reward {sent by} the central node is negative.

\section{Random access with Q-Learning}\label{sec:qlearning}

Q-Learning is a type of machine learning (ML), which is model-free and it can be implemented in a distributed way and with low complexity. The advantage of using Q-Learning to solve the RA problem is that it is easily implemented on thousands of mMTC devices due to its low complexity, {while the devices decide in a distributed and decentralized way} the best time-slots to transmit based on previous transmissions. The learning method of each device can be modeled as a Markov decision process (MDP), where the change to a future state depends on the factors: the current state, the transition probability function and the {reward value} \cite{Barto_RL}.

The $n$-th device has a Q-value, {namely} $Q_{n,k}^t$, that indicates the preference to transmit in the $k$-th time-slot {and} step $t$. All Q-values make up a Q-table of $N$ rows and $K$ columns. Initially, the entire Q-table is {set} to zero: $Q_{n,k}=0, \forall n \in 1,\dots, N, \; \forall k \in 1, \dots, K$. {Hence,
 in} order to transmit a packet, the device selects the time-slot with the highest Q-value from its Q-table. If there is more than one time-slot whose Q-value is the maximum, then the choice is random among these Q-values.

At the end of the frame, the central node sends a reward to each device indicating whether the transmission was successful or not in a given time-slot. Thus the Q-value in the next step of the $n$-th device and $k$-th time-slot is updated to
\begin{equation}\label{eq:Q_update}
 Q_{n,k}^{t+1} = Q_{n,k}^t + \alpha(R_{n,k} - Q_{n,k}^t)
\end{equation}
where $R_{n,k}$ is the reward transmitted by central node, and $\alpha$ is the learning rate. The learning rate is a weight {value in the range $\alpha\in[0;\,1]$. In this work, $\alpha$ is assumed fixed and equal for all devices in the system.}

The {Q-table update, and the packet transmission are performed subsequently} until {each} device transmits all of its packets. The {reward-based RA} algorithm is considered to have converged when all devices have transmitted all their packets. In the convergence {process}, it is defined that the total number of successes is $S$, the total number of failures is $F$ and the total number of time-slots spent is $T$.

\subsection{Independent Q-Learning}

The independent Q-Learning technique requires that central node send only one bit (b = 1) for each device. The reward sent to the $n$-th device that chose the $k$-th time-slot {is simply defined as}: \cite{Sharma2019_CongestionMinimization}:
\begin{equation}
R_{n,k}^{\textsc{ind}} = \begin{cases}
    +1, \;\; \text{if transmission succeeds,}\\
    -1, \;\; \text{otherwise.}
    \end{cases}
\end{equation}
Therefore, if only the $n$-th device has chosen the $k$-th slot, the transmission is successful and the reward is +1. If two or more devices choose the $k$-th slot, a collision occurs and the reward is -1 for all of them. {Reward} $R_{n,k}^{\textsc{ind}}$ is used to update the Q-table for all devices and time-slots through Eq. \eqref{eq:Q_update}.

\subsection{Collaborative Q-Learning}
Assuming that central node is aware of the number of devices that tried to access the $k$-th slot, it is possible to define a congestion level $C_k$ in slot $k$, given by
\begin{equation}
C_k = \dfrac{|\psi_k|}{N},
\end{equation}
As a result, the reward sent by central node to the $n$-th device that chose the $k$-th time-slot is given by \cite{Sharma2019_CongestionMinimization}
\begin{equation}
R_{n,k}^{\textsc{col}} = \begin{cases}
+1, \;\; \text{if transmission succeeds,}\\
-{\mathcal{M}_b}\{C_k\}, \;\; \text{otherwise,}
\end{cases}
\end{equation}
where ${\mathcal{M}_b}\{C_k\}$ is a quantized value of $C_k$ based on the number of bits $b$ available for the header, {e.g.}, if $b$ = 2 bits and {assuming} that the level of congestion varies from 0 to 1, then {the reward values can be  unambiguously represented by four} quantized levels, ${\mathcal{M}_b}\{C_k\} \in \{0.25, 0.5, 0.75, 1\}$.

As $C_k$ in this case is a real number, then the central node should transmit a quantized version of such real number, decreasing the spectral efficiency of transmission, and the devices will have to use a certain number of quantization bits $b$ to represent this real value. Therefore, there is a trade-off between bandwidth overhead and accuracy when quantized version (limited number of bits)  is transmitted by the central node and the true value of the reward. Hence, the fixed number of bit of quantization must be selected carefully.

The advantage of the collaborative method over the independent one is that the devices learn to choose the time-slots with lower levels of congestion to transmit their packages. The disadvantage is that the central node needs to know the number of interfering devices. In addition, the reward becomes a real number and no longer a bit, as in the independent method.

\subsection{Distributed Packet-based {RA} {for Crowded MTC Scenarios}}

With the increase in the number of devices and the increase in the probability of collision in crowded mMTC mode, it becomes more difficult for central node to identify the number of interfering devices. Therefore, the advantage {of the collaborative Q-learning technique in regions with high density} of devices depends on an ideal non-feasible scenario. In addition, independent and collaborative Q-learning techniques are not completely fair, as a time-slot becomes unique for one device over the entire learning period, while the other devices continue to collide and expect to randomly find a suitable time-slot to finish transmitting all packets.

Therefore, this work proposes a distributed packet-based Q-Learning random access technique where {the Q-table updating} takes into account the number of remaining {packet} that each device still has to transmit in that frame. The higher this number, the greater is the {respective} reward{, increasing the frequency of transmission attempt in that time-slot}; hence, it is expected that on average all devices finish transmitting their packets at the same time.

Let's define the factor $\epsilon_n$ {for each device} as:
\begin{equation}
    \epsilon_n = 1 - \dfrac{\ell_n}{L},
\end{equation}
where $\ell_n$ is the number of remaining packets to be transmitted by the $n$-th device; {hence,} when the device has already transmitted a large number of packets, $\epsilon_n$ tends to 1.

{In the proposed Q-learning-based RA method, the} reward sent by central node to the $n$-th device at the $k$-th time-slot {is defined in a same way as in the independent Q-learning method}: 
\begin{equation}\label{eq:Rdistr}
R_{n,k}^{\textsc{pac}} = R_{n,k}^{\textsc{ind}} = \begin{cases}
   +1, \;\; \text{if transmission succeeds,}\\
    -1, \;\; \text{otherwise.}
    \end{cases}
\end{equation}
However, {since the proposed method is totally distributed, the reward processing is utterly done} by the devices. {Hence, under this method, the Q-Table updating} takes into account the number of packets that the device still has to transmit {results}:
\begin{align}
    Q_{n,k}^{t+1} & = \begin{cases}\label{eq:Qdistr}
        Q_{n,k}^t + \alpha(R_{n,k}^{\textsc{pac}} - Q_{n,k}^t), \;\; \text{if Tx succeeds,}\\
        Q_{n,k}^t + \alpha(\epsilon_n R_{n,k}^{\textsc{pac}} - Q_{n,k}^t), \;\; \text{otherwise.}
    \end{cases}\\
    & = \begin{cases}\label{eq:Qdistr_sub}
        Q_{n,k}^t + \alpha(1 - Q_{n,k}^t), \;\; \text{if Tx succeeds,}\\
        Q_{n,k}^t - \alpha(\epsilon_n + Q_{n,k}^t), \;\; \text{otherwise.}
    \end{cases}
\end{align}
{where eq. \eqref{eq:Qdistr_sub} can be obtained by substituting \eqref{eq:Rdistr} into \eqref{eq:Qdistr}.}

{Notice that in} the distributed packet-based RA method, the central node does not need to know the number of devices that has collided in a given time-slot. Therefore, the reward to be transmitted is binary ($b$ = 1), {requiring the same infra-structure than} the independent technique. In addition, among the devices that collided, devices that need to transmit more packets are privileged with a more positive reward compared to {those devices with less packets remaining to be transmitted}, making the technique more {appropriate to attain improved throughput-complexity tradeoff when compared to collaborative and independent-like methods}. The pseudo code for the proposed distributed packet-based technique is present in Algorithm \ref{algoritmo}.

\begin{algorithm}[!htbp]
	\caption{\bf Distributed Packet-Based RA Method}
	\label{algoritmo}
	\begin{algorithmic}
		\State Initialize $Q_{n,k} = 0,\; \forall n \in {\mathcal{N}}, \; \forall k \in {\mathcal{K}}$
		\State Initialize $\ell_n = L, \; \forall n \in \mathcal{N}$; \quad 
		$T = 0, \; S = 0$
		\While{$\sum_{n=1}^{N} \ell_{n} > 0$}
		    \State Initialize $c_n = 0, \; \forall n \in \mathcal{N}$
		    \For{$n = 1:N$}
		        \If{$\ell_n > 0$}
		            \State $\mathcal{C}_n = \{k \in \mathcal{K} \; | \; Q_{n,k} = \max\limits_{k}\{Q_{n,k}\} \}$
		            \State Select randomly: $c_n \in \mathcal{C}_n$
	            \EndIf
	        \EndFor
		    \For{$k = 1:K$}
		        \State $T \leftarrow T + 1$
		        \State $\psi_k = \{n \in \mathcal{N} \; | \; c_n = k\}$
    		    \If{$|\psi_k| = 1$}
    		  \State $S \leftarrow S + 1$
    		        \State ${R_{n,k}^{\textsc{pac}} = +1, \; \forall n \in \psi_k}$
    		        \State {$Q_{n,k} \leftarrow Q_{n,k} + \alpha(\ta{1} - Q_{n,k}), \; \forall n \in \psi_k$}
    		        \State $\ell_n \leftarrow \ell_n - 1, \; \forall n \in \psi_k$
    		    \ElsIf{$|\psi_k| > 1$}
    		        \State $\epsilon_n = 1 - \frac{\ell_n}{L}, \; \forall n \in \psi_k$
    		        \State $R_{n,k}^{\textsc{pac}} = -1, \; \forall n \in \psi_k$
    		        \State $Q_{n,k} \leftarrow Q_{n,k} \ta{-} \alpha(\epsilon_n \ta{+} Q_{n,k}), \; \forall n \in \psi_k $
    		    \EndIf
		    \EndFor
		\EndWhile
	\end{algorithmic}
\end{algorithm}

\section{Numerical results}\label{sec:results}
In this section, {proposed} Q-Learning RA technique is {numerically validated via computer simulations, and compared  with the independent and collaborative learning methods.} In order to guarantee an average behavior of the number of transmissions carried out successfully, {$10^4$} realizations for each experiment were considered. The main simulation parameter values used are shown in Table \ref{tab:parameters}.

\begin{table}[!htbp]
    \centering
    \caption{Numerical parameters.}
    \label{tab:parameters}
    \begin{tabular}{rl}
        \hline
        \textbf{Parameter} & \textbf{Value} \\
        \hline
        Monte-Carlo realizations & $N_{\text{reps}}$ = 10,000 \\
        Time-slots per frame & $K$ = 400 \\
        Network Loading factor & $\mathcal{L} = \frac{N}{K} \in [0.25;\,\, 3.00]$ \\
        Packets per device & $L \in [50; 500]$ \\
        Learning rate & $\alpha \in [0.05; 0.5]$ \\
        \hline
        Header bits ({collab.}) & $b\in[1; 2; 4; 8; 16]$ bits\\ 
        Payload bits & $p\in[1; 2; 4; 8; \ldots ; 256]$ bits\\
        \hline
    \end{tabular}
\end{table}

{An important figure of merit is the normalized throughput, defined as} the ratio between the number of successful {packet} transmissions{, $S$,} and the {corresponding} number of time-slots required{, $T$}. However, as not all bits in the transmission are data from the devices, {so the ratio between the payload bits and reward bits should be} taken into account. Hence, {the normalized} throughput is defined as
\begin{equation}
   {\mathcal{T}} = \left(\dfrac{p}{b + p}\right) \dfrac{S}{T} = \left(\dfrac{p}{b + p}\right) \dfrac{N L}{T}.
\end{equation}
The calculation of normalized throughput is performed after the convergence of the algorithm, when all devices transmit all their packets, and it indicates how {efficiently the time-frames have being used in each RA method.}

\subsection{Number of bits of quantized collaborative reward}

\begin{figure}[!htbp]
    \centering
    \includegraphics[trim=7mm 2mm 16mm 14mm, clip,width=1\columnwidth]{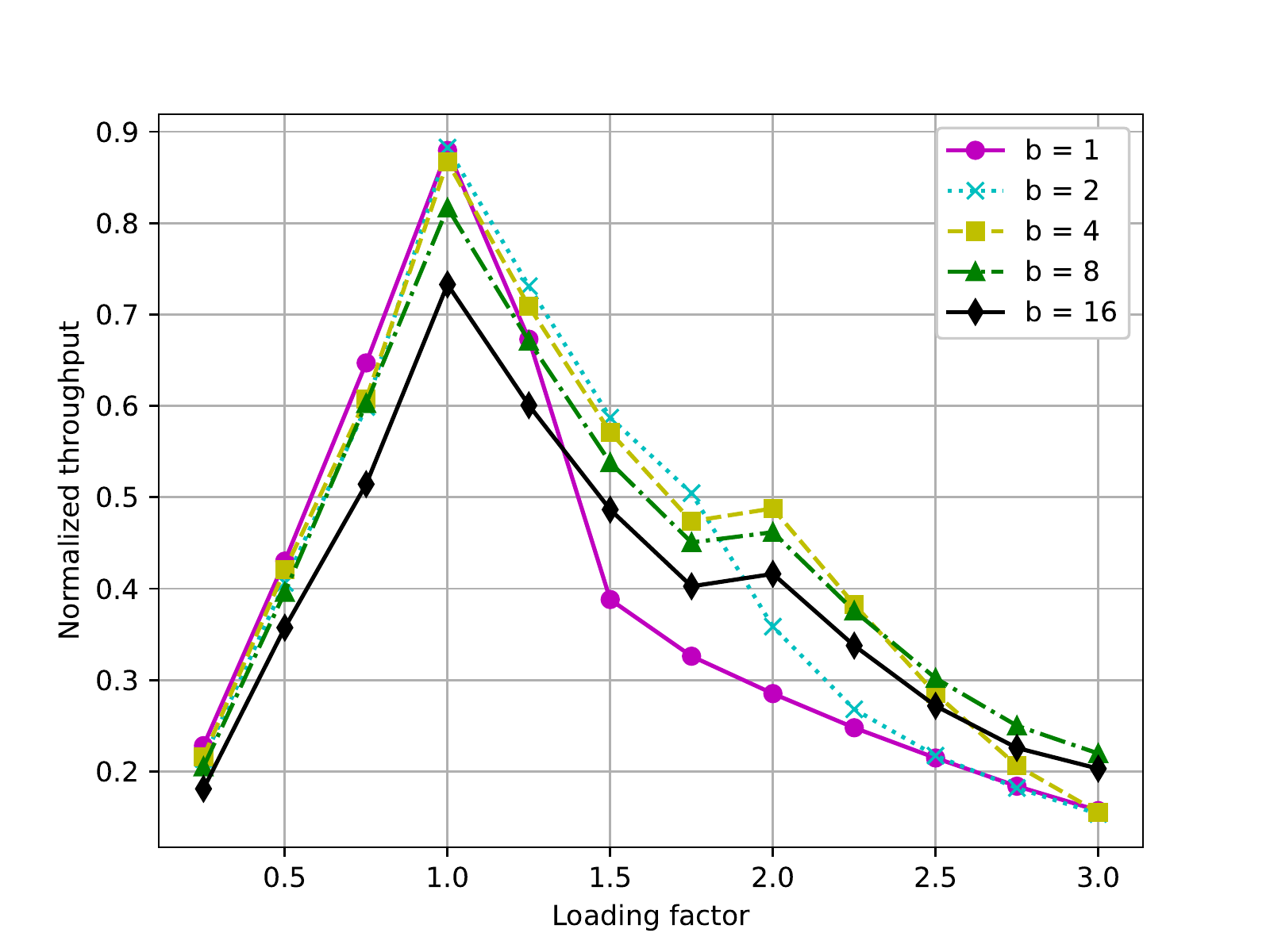}
    \vspace{-6mm}
    \caption{Throughput for collaborative method varying the loading factor, considering $p$ = 64, $L$ = 100, and $\alpha$ = 0.1.}
    \label{fig:nbits}
\end{figure}

To find the smallest number of bits that results in a suitable accuracy in representing the actual number of the congestion level in the collaborative technique without reducing the throughput, Fig. \ref{fig:nbits} depicts the average throughput calculated as a function of the loading factor $\mathcal{L}$. The result shows that, within the analyzed scenario, a {suitable tradeoff} choice for the number of quantization bits that maximize the mean throughput in the collaborative Q-Learning technique is $b=4$ bits. {By deploying four bits, it is possible to attain} a good level of quantization for the real number of the reward, but without reducing the throughput due to the increase in header bits{; hereafter,} this value was adopted in all simulations of the collaborative technique.

\subsection{Normalized Throughput}

\begin{figure}[!htbp]
\centering
\includegraphics[trim=7mm 2mm 16mm 14mm, clip,width=1\columnwidth]{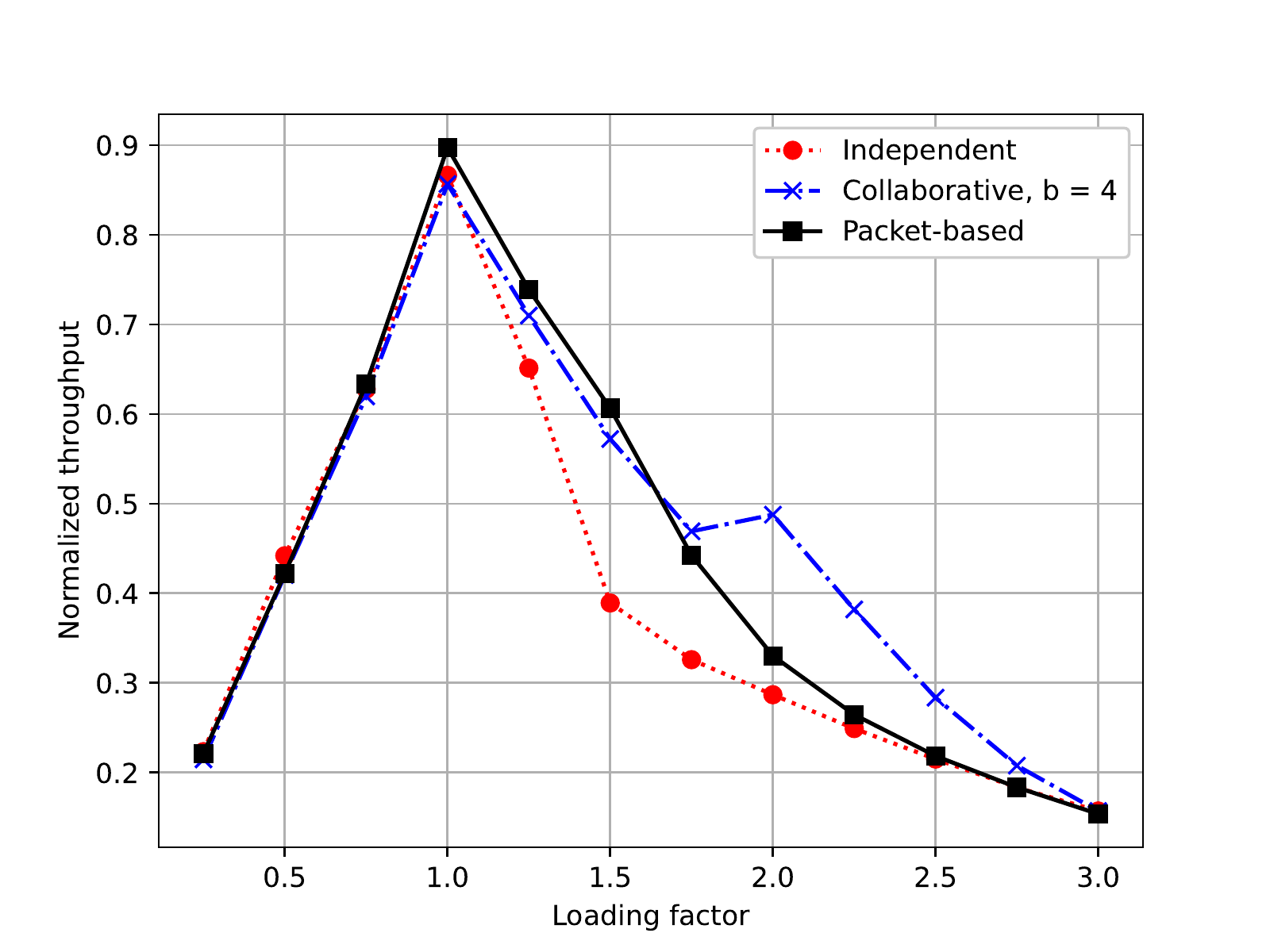}
\vspace{-6mm}
    \caption{Normalized throughput in function of loading factor for independent, collaborative, and packet-based Q-Learning, considering $p$ = 64, $L$ = 100, $K$ = 400, and $\alpha$ = 0.1.}
    \label{fig:throughput_devices}
\end{figure}

In Fig. \ref{fig:throughput_devices}, the throughput is analyzed as a function of the loading factor. It is observed that the maximum throughput is obtained when $\mathcal{L} = 1$ for all techniques, because in this scenario, the frame is being used with greater efficiency, where in average there is a time-slot for each device. {Hence, as expected, the throughput is lower in underloaded and overloaded scenarios, {\it i.e.,} $\mathcal{L} \neq 1$}, where there are fewer {or many} devices than time-slots and is not the ideal scenario, as more and more devices could be allocated on the network to increase {the spectral} efficiency. {In particular, we are interest in crowded MTC scenarios.}

In the $\mathcal{L} > 1$ scenario, the RA techniques start to have a worse throughput because, when there are more devices than time-slots, the probability of collision increases {substantially and unavoidably}, which consequently reduces the success probability and throughput. The difference between independent and collaborative {Q-Learning RA} techniques stands out in this {important} scenario {of practical interest}. The collaborative technique has greater throughput because the central node indicates to the devices which time-slots have the highest congestion level, through the information sent as a reward. The devices learn to transmit in the least congested time-slots, thus reducing the probability of collision and, consequently, increasing throughput.

The performance of the packet-based technique is superior to other techniques up to $\mathcal{L}$ = 1.6. From that point on, the collaborative technique becomes superior {in the interval $1.6\leq \mathcal{L} \leq 3.0$,} and then the techniques converge to the same throughput value. It is expected that the collaborative technique presents a higher throughput in relation to the others in the medium-high congestion scenarios ($1.75 \leq \mathcal{L} \leq 3.0$) because the reward sent by the central node provides more details about the level of congestion of each time-slot. However, the packet-based technique still proves to be superior to the independent one in this scenario, in addition to being less complex than the collaborative one in relation to the central node, since the reward sent is binary.

\subsection{{Asymptotic Throughput}}

\begin{figure}[!htb]
    \centering
    \includegraphics[trim=2mm 2mm 16mm 14mm, clip,width=1\columnwidth]{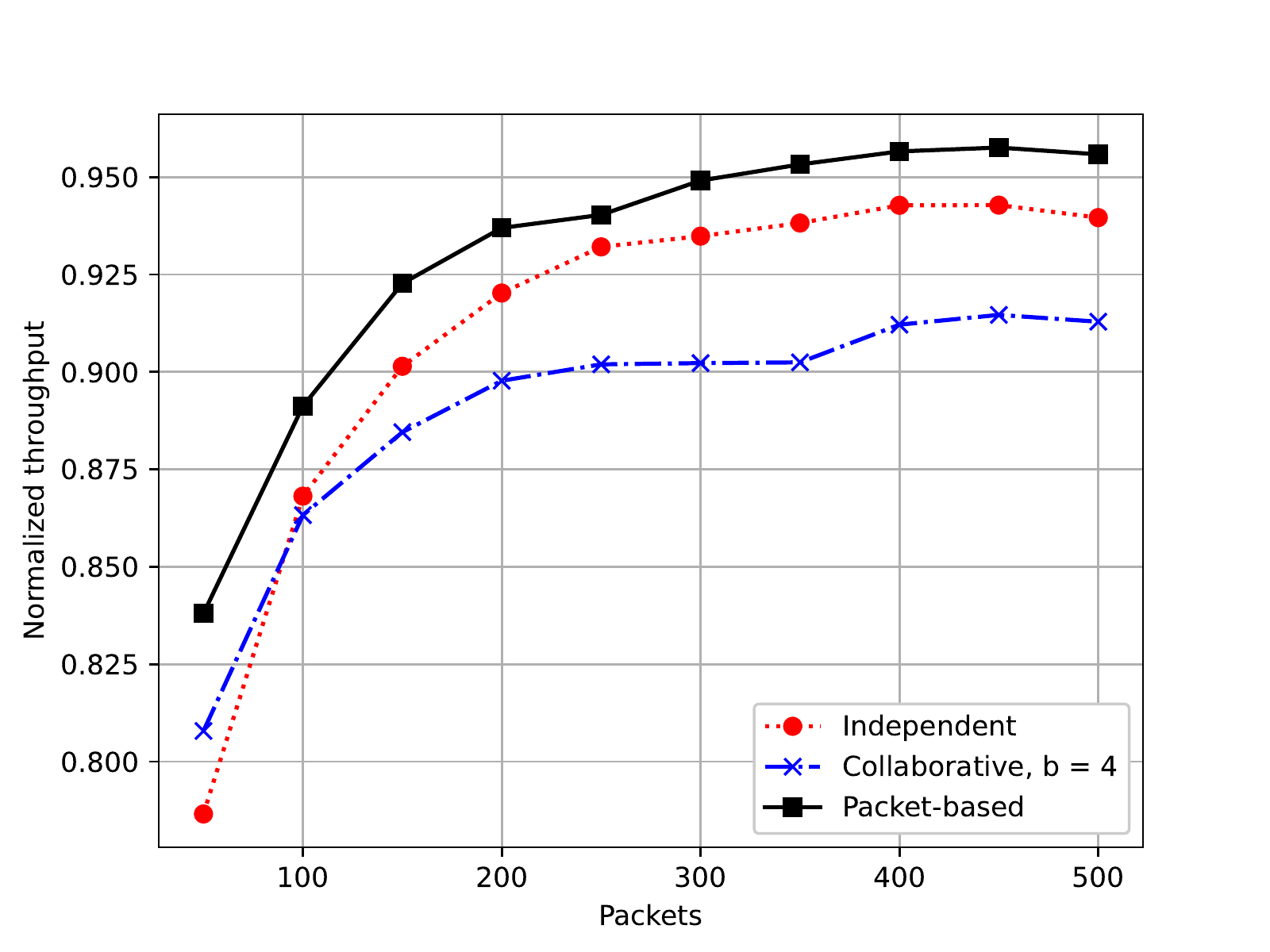}
    \vspace{-6mm}
    \caption{Throughput as a function of the number of packets, considering loading factor $\mathcal{L}$ = 1, $K$ = 400 time-slots, $p$ = 64 bits, and $\alpha$ = 0.1.}
    \label{fig:packets}
\end{figure}

In Fig. \ref{fig:packets}, the throughput for the three reward techniques was analyzed with the change in the number of packets that each device has to transmit, from $L$ = 50 to $L$ = 500 packets. For this result, we consider $\mathcal{L} = 1$, {payload $p=64$ bits,} and learning rate $\alpha = 0.1$.

It is possible to conclude that the throughput increases with the increase in the number of packets. This is because the number of successes increases, without having a significant increase in the number of time-slots needed to transmit all packets. However, the curves begin to converge to a constant value. This indicates that, even if the number of packets increases, the time to transmit them in the same proportion is increased, which makes the throughput constant. {The proposed distributed packet-based RA method reveals a superior {\it asymptotic normalized throughput}:
$$
\mathcal{T}_{\infty}(\mathcal{L}) = \lim_{L, T\rightarrow \infty} \left(\dfrac{p}{b + p}\right) \dfrac{N L}{T},
$$
resulting for the specific network loading factor: $\mathcal{T}_{\infty}^{\textsc{pac}}(1)\approx 0.965$; \,\, $\mathcal{T}_{\infty}^{\textsc{ind}}(1) \approx 0.940$; \,\, $\mathcal{T}_{\infty}^{\textsc{col}}(1) \approx 0.915$
}

\subsection{Payload bits}
\begin{figure}[!htb]
    \centering
    \includegraphics[trim=7mm 2mm 16mm 14mm, clip,width=1\columnwidth]{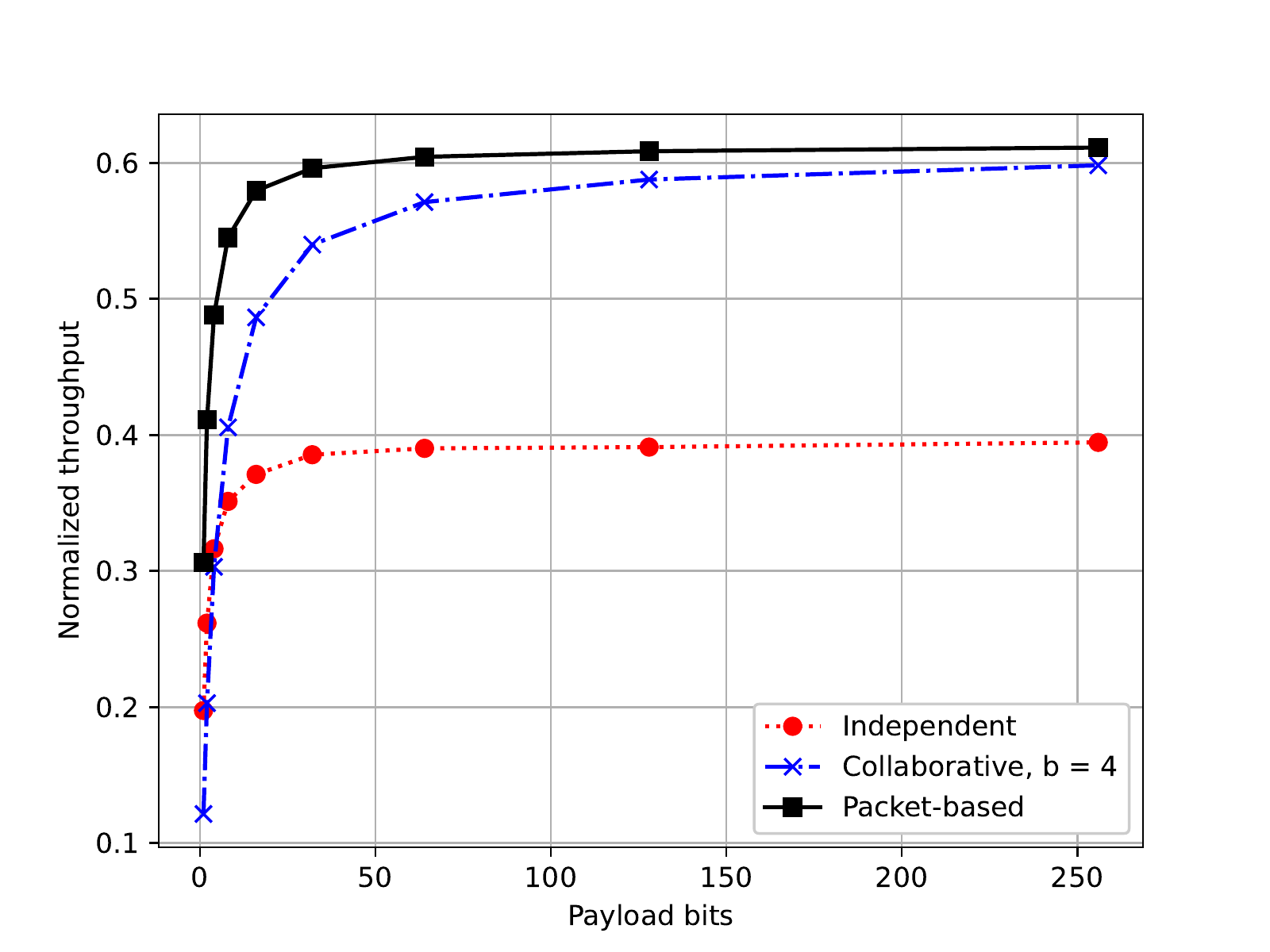}
    \vspace{-5mm}
    \caption{Normalized throughput as as function of payload bits, considering $\mathcal{L}$ = 1.5, $\alpha$ = 0.1, and $L$ = 100.}
    \label{fig:payload_bits}
\end{figure}

Fig. \ref{fig:payload_bits} shows the result of the throughput as a function of the number of payload bits $b$. The numerical result indicates that when the number of payload bits is small, with a value close to the number of header bits, the throughput is low. As the number of payload bits increases, the throughput increases until it converges to a ceiling value. This convergence occurs in our configuration setup close to $p$ = 64, and for this reason, this payload value was considered in the rest of the simulations in this work.

As the collaborative technique has a larger number of header bits ($b$ = 4), then it depends on a larger number of payload bits to present the same throughput as the packet-based technique. For example, to achieve a normalized throughput of $\mathcal{T}_p = 0.5$, the collaborative technique needs 16 bits of payload, while the packet-based one needs 4 bits in the analyzed scenario. The reduction in the number of payload bits can be an advantage in simplifying the process {in which a bunch of devices randomly access the channel and transmitting their packets}.

\subsection{Latency}

\begin{figure}[!htbp]
    \centering
    \includegraphics[trim=11mm 3mm 16mm 11mm, clip, width=1\columnwidth]{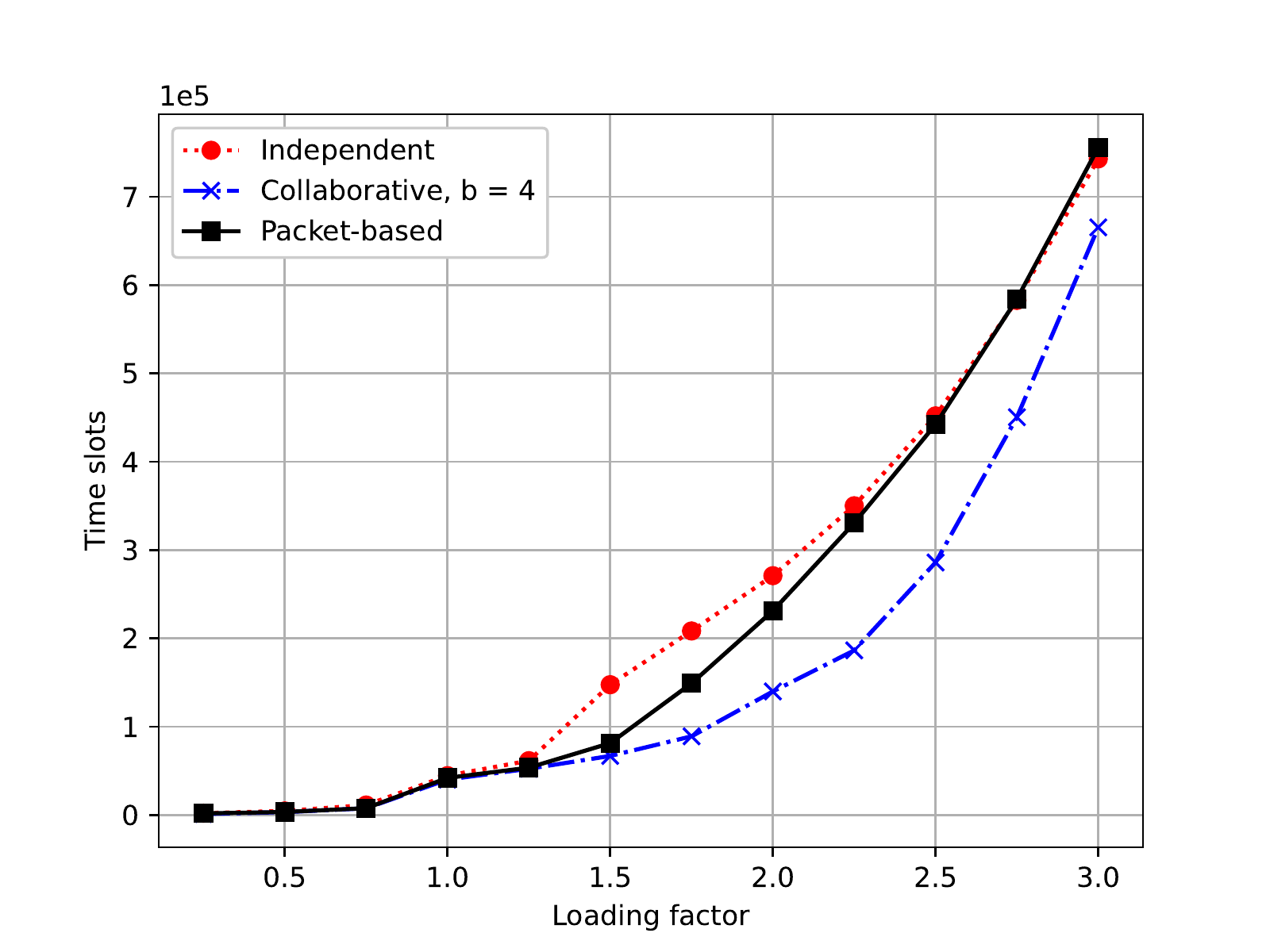}
    \vspace{-6mm}
    \caption{Total number of time-slots as a function of loading factor considering $L$ = 100, and $\alpha$ = 0.1.}
    \label{fig:timeslots}
\end{figure}

Latency in this work is defined as the total amount of time-slots $T$ that all devices need to transmit {a fixed number of} packets. In Fig. \ref{fig:timeslots}, there is an analysis of the total number of time-slots required for the complete transmission of {$L=100$ packets/device} according to {an increasing in the} loading factor of the system.

{As expected, latency in terms of total time-slots required} increases when the loading factor increases, as the system becomes more congested and the number of collisions increases, requiring more retransmissions. The result can be analyzed in three different scenarios. For a {low-medium loading factor, $\mathcal{L}\leq 1.0$,} all techniques have the same latency. {For slightly-crowded and crowded scenarios, {\it i.e.,} $1.2\leq \mathcal{L}\leq 2.5$,}, the independent technique has the highest latency in relation to the others. Finally, for {high-loading over-crowded scenarios,} the Q-Learning packet-based RA technique approaches the latency of the independent technique, while the collaborative technique {holds} the lowest latency. 

Hence, from Fig. \ref{fig:throughput_devices}, \ref{fig:packets}, {\ref{fig:payload_bits}, and \ref{fig:timeslots},} one can infer that the proposed distributed packet-based RA method attains the best throughput-latency trade-off for a wide range loading factors, $0.75 \leq \mathcal{L}\leq  2.5${, mainly in typically (over)crowded scenarios}.

\subsection{Learning Rate}
\begin{figure}[!htb]
 \centering
 \includegraphics[trim=7mm 2mm 16mm 9mm, clip, clip,width=1\columnwidth]{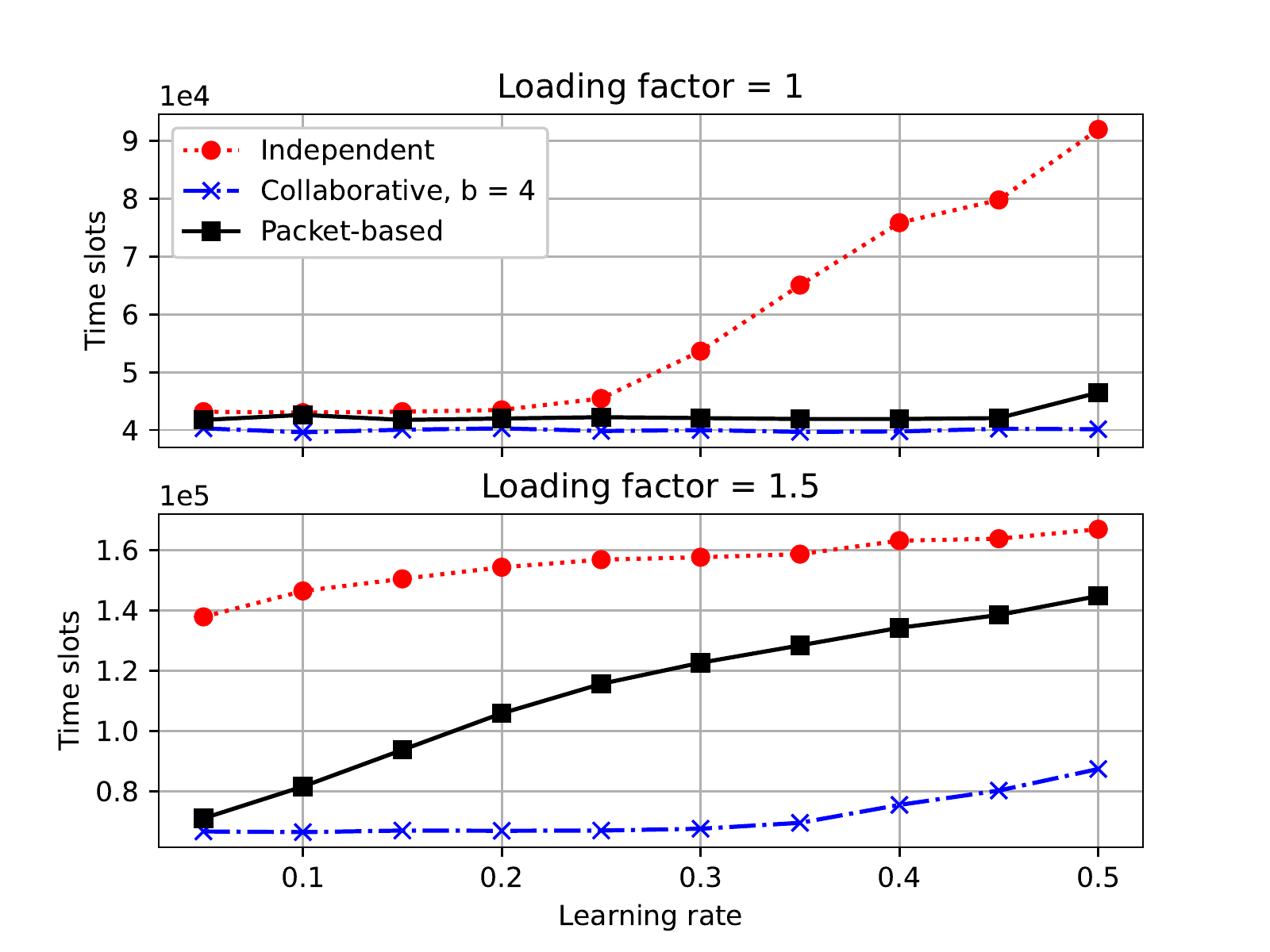}
\vspace{-6mm}
 \caption{Total number of time-slots as a function of learning rate considering $L$ = 100.}
    \label{fig:learningrate}
\end{figure}

Finally, {the adopted value for the learning rate is justified. For that, latency was evaluated according to the learning rate, as shown in Fig.} \ref{fig:learningrate}. The {adoption of an increasing value for} the learning rate negatively affects the performance of {reward-based RA} techniques {in crowded scenarios}, as the latency to achieve convergence increases. When the learning rate is high, the weight given to the reward of the central node is greater. {Hence,} in more congested scenarios{, more negative than positive rewards can be expected}. Therefore, when devices give greater weight to negative rewards, the latency of the technique increases. This behavior is observed when $\mathcal{L}$ = 1.5, as the latency increases significantly with the increase in the learning rate. When $\mathcal{L}$ = 1, this behavior is smoothed, since the increase in latency only occurs when $\alpha = 0.5$ for the collaborative and packet-based techniques.

\section{Conclusions}\label{sec:conclusions}
Q-Learning-based random access methods for mMTC networks have been investigated in terms of throughput and latency. The numerical results and analyses have demonstrated that the proposed distributed packet-based RA method attains higher throughput combined with lower latency than the conventional independent Q-learning RA technique, even with the central node transmitting only a bit of reward for both existing techniques. In addition, the proposed distributed packet-based method conferred the best throughput-latency trade-off regarding both existing techniques for different loading factor scenarios ($0.25 \leq \mathcal{L} \leq 1.5$). Such trade-off improvement reduces the number of bits transmitted by the central node to one while distributing the processing among the devices. Finally, in highly congested machine type scenarios, {\it e.g.}, $\mathcal{L} \approx 3$, the throughput of the proposed technique is the same as that of the collaborative technique.

\section*{Acknowledgements}
This study was supported in part by the CAPES Scholarship (first author); in part it was supported by the National Council for Scientific and Technological Development (CNPq) of Brazil under Grants 310681/2019-7, and in part by Londrina State University, Paran\'a State Government (UEL).\\
	


\end{document}